\pdfoutput=1
\documentclass{article} 
\usepackage{iclr2018_conference,times}
\usepackage{hyperref}
\usepackage{url}
\usepackage{amsmath}
\usepackage{amssymb}
\usepackage{algorithm}
\usepackage{algpseudocode}
\usepackage{algorithmicx}
\usepackage{booktabs}
\usepackage{comment}
\usepackage{tikz}
\usepackage{asypictureB}
\usepackage{bm}
\usepackage{caption}
\usetikzlibrary{backgrounds}
\usetikzlibrary{calc}
\usetikzlibrary{fit}
\usetikzlibrary{positioning}
\usepackage{enumitem}
\usepackage{times}
\usepackage{graphicx} 
\usepackage{subfigure} 
\usepackage{tikz}
\usetikzlibrary{backgrounds}
\usetikzlibrary{calc}
\usetikzlibrary{fit}
\usetikzlibrary{positioning}
\usepackage{lipsum,adjustbox}
\usepackage{amsmath}
\usepackage{bm}

\usepackage{amssymb,amsthm,amsfonts,amscd} 
\usepackage{mathrsfs}
\usepackage{mathtools}
\usepackage{tabularx}
\usepackage{lipsum}
\usepackage{cellspace} 
\usepackage{makecell} 
\setcellgapes{5pt}

\usepackage{hyperref}
\usepackage[acronym,smallcaps,nowarn]{glossaries}
\newacronym{ACtuAL}{actual}{actor critic under adversarial learning}
\definecolor{mygreen}{HTML}{167dde}
\definecolor{myred}{HTML}{f22835}

\colorlet{greenfill}{mygreen!20!white}
\colorlet{redfill}{myred!20!white}
\colorlet{moreredfill}{myred!40!white}

\newcommand{\EE}{\mathbb{E}}
\newcommand{\RR}{\mathbb{R}}

\DeclareMathOperator*{\argmax}{\arg\max}
\DeclareMathOperator*{\argmin}{\arg\min}

\title{ACtuAL: Actor-Critic under Adversarial Learning}

\author{Anirudh  Goyal$^{1}$,  {\bf Nan Rosemary Ke$^{12}$}, {\bf Alex Lamb$^{1}$},  {\bf R Devon Hjelm$^1$} \\
{\bf Chris Pal$^{12}$},  {\bf Joelle Pineau$^{13}$}, {\bf Yoshua Bengio$^{1\dagger}$}\\
$^1$ MILA, Universit\'e de Montr\'eal\\
$^2$ Polytechnique Montr\'eal \\ 
$^3$ McGill University \\
 $^\dagger$CIFAR Senior Fellow\\
}

\iclrfinalcopy
\begin{document}

\maketitle
\begin{abstract}

Generative Adversarial Networks (GANs) are a powerful framework for deep generative modeling.  Posed as a two-player minimax problem, GANs are typically trained end-to-end on real-valued data and can be used to train a generator of high-dimensional and realistic images.  
However, a major limitation of GANs is that training relies on passing gradients from the discriminator through the generator via back-propagation. 
This makes it fundamentally difficult to train GANs with discrete data, as generation in this case typically involves a non-differentiable function.  
These difficulties extend to the reinforcement learning setting when the action space is composed of discrete decisions.
We address these issues by reframing the GAN framework so that the generator is no longer trained using gradients through the discriminator, but is instead trained using a learned critic in the actor-critic framework with a Temporal Difference (TD) objective.   
This is a natural fit for sequence modeling and we use it to achieve improvements on language modeling tasks over the standard Teacher-Forcing methods.
\end{abstract}

\vspace{-2mm}
\section{Introduction}
\vspace{-2mm}
Sequence generation is an important and broad family of machine learning problems that includes image generation, language generation, and speech synthesis. 
Variants of recurrent neural networks~\citep[RNNs,][]{hochreiter1997long, chung2014empirical, cho2014learning} have attained state-of-the-art for many sequence to sequence tasks such as language modeling~\citep{mikolov2012context}, machine translation~\citep{sutskever2014sequence}, speech recognition~\citep{chorowski2015attention} and time series forecasting~\citep{flunkert2017deepar}.
Much of RNN's popularity is derived from its ability to handle variable-length input and output along with a simple learning algorithm based on back-propagation through time (BPTT).

Normally, generative RNNs are trained to maximize the likelihood of samples from an empirical distribution of target sequences, i.e., using maximum-likelihood estimation~(MLE), which essentially minimizes the KL-divergence between the distribution of target sequences and the distribution defined by the model.
While principled and effective, this KL-divergence objective tends to favor a model that overestimates its uncertainty / smoothness, which can lead to unrealistic samples~\citep{goodfellow2016nips}.

For a purely generative RNN, the desired behavior is for the output from the free-running sampling process to match the samples in the target distribution.
The most common approach to training RNNs is to maximize the likelihood of each token from the target sequences given previous tokens in the same sequence (a.k.a., Teacher Forcing~\citep{williams1989learning}).
In principle, if the generated sequences match the target sequences perfectly, then then generative model is the same as the one from the training procedure.
In practice, however, because of the directed dependence structure in RNNs, small deviations from the target sequence can cause a large discrepancy between the training and evaluation model, so it is typical to employ
beam search or scheduled sampling to incorporate the real generative model during training~\citep{bengio2015scheduled}.

Generative adversarial networks~\citep[GANs][]{goodfellow2014generative} estimate a \emph{difference measure}~\citep[i.e., a divergence, e.g., the KL or Jensen-Shannon or a distance, e.g., the Wasserstein][]{nowozin2016f, arjovsky2017wgan} using a binary classifier called a \emph{discriminator} trained only on samples from the target and generated distributions.
GANs rely on back-propagating this difference estimate through the generated samples to train the generator to minimize the estimated difference.

In many important applications in NLP, however, sequences are composed of discrete elements, such as character- or word-based representations of natural language, and exact back-propagation through generated samples is not possible.
There are many approximations to the back-propagated gradient through discrete variables, but in general credit assignment with discrete variables is an active and unsolved area of research~\citep{bengio2013estimating, gu2015muprop, maddison2016concrete, jang2016categorical, tucker2017rebar}. 
A direct solution was proposed in boundary-seeking GANs~\citep[BGANs][]{hjelm2017bsgan}, but the method does not address difficulties associated with credit assignment across a generated sequence.

In this work, we propose Actor-Critic under Adversarial Learning (ACtuAL), to overcome limitations in existing generative RNNs. 
Our model is inspired by the classical actor-critic algorithms in reinforcement learning, and we extend this algorithm in the setting of sequence generation to incorporate a discriminator. 
Here, we treat the generator as the ``actor", whose actions are evaluated by a ``critic" that estimates a reward signal provided by the discriminator. 
Our newly proposed ACtuAL framework offers the following contributions: 
\begin{itemize}
\item A novel way of applying Temporal Difference (TD) learning to capture long-term dependencies in generative models without full back-propagation through time.  
\item A new method for training generative models on discrete sequences.
\item A novel regularizer for RNNs, which improves generalization in terms of likelihood, on both character- and word-level language modeling with several language model benchmarks.

\end{itemize}

\section{Background}

\subsection{Generative RNNs and Teacher Forcing}
A generative recurrent neural networks (RNN) models a joint distribution of a sequence, $Y = y_1, y_2, \dots, y_T$, as a product of conditional probability distributions,
\begin{align}
    q(y_1, y_2, \dots, y_T) = q(y_1) \prod_{t=2}^T q(y_t \mid y_1 \dots y_{t-1}). \nonumber
\end{align}
RNNs are commonly trained with Teacher Forcing~\citep{williams1989learning}, where at each time step, $t$, of training, the model is judged based on the likelihood of the target, $y_t$, given the ground-truth sequence, $y_1 \dots, y_{t-1}$.

In principle, RNNs trained by teacher forcing should be able to model a distribution that matches the target distribution, as the joint distribution will be modeled perfectly if the RNN models each one-step-ahead prediction perfectly. 
However, if the training procedure is not able to model the conditional  distribution perfectly, during the \emph{free-running} phase, even small deviations from the ground-truth may push the model into a regime it has never seen during training, and the model may not able to recover from these mistakes.
From this, it is clear that the optimal model under Teacher Forcing is undesirable, as we want our generative model to generalize well.
This problem can be addressed with beam-search or scheduled sampling~\citep{bengio2015scheduled}, which can be used to add some exploration to the training procedure.
However, the central limitation is in these approaches is that training with MLE necessitates that the model stays close to the target distribution, which can limit overall capacity and generalization properties.

\subsection{Generative Adversarial Networks}
Generative Adversarial Networks~\citep[GANs,][]{goodfellow2014generative} define a framework for training a generative model by posing it as a minimax game. 
GANs have been shown to generate realistic-looking images that are more crisp than the models trained under maximum likelihood~\citep{radford2015unsupervised}. 

Let us assume that we are given empirical samples from a target distribution over a domain $\mathcal{X}$ (such as $\RR^n$), $p(x)$.
In its original, canonical form, a GAN is composed of two parts, a generator network, $G_{\theta}: \mathcal{Z} \rightarrow \mathcal{X}$, and a discriminator, $D_{\phi}: \mathcal{X} \rightarrow \RR$. The generator takes as input random noise, $z \in \mathcal{Z}$ from a simple prior $p(z)$ (such as a spherical Gaussian or uniform distribution), and produces a generated distribution, $q_{\theta} (x)$ over $\mathcal{X}$.
The discriminator then is optimized to minimize the mis-classification rate, while the generator is optimized to maximize it:
\begin{align}
    (\hat{\theta}, \hat{\phi}) = \argmin_\theta \argmax_{\phi} \EE_{p(x)}[\log{D_{\phi}(x)}] + \EE_{q_{\theta}(x)}[\log{(1 - D_{\phi}(x))}]
    \nonumber\\
    = \argmin_\theta \argmax_{\phi} \EE_{p(x)}[\log{D_{\phi}(x)}] + \EE_{p(z)}[\log{(1 - D_{\phi}(G_{\theta}(z)))}]
    \label{eq:gan}
\end{align}

As the discriminator improves, it better estimates $2 * JSD(p, q_{\theta}) + \log{4}$, where $JSD$ is the Jensen-Shannon divergence.
The advantage of this approach is that, as the discriminator provides an estimate of this divergence based purely on samples from the generator and the target distribution, unlike Teacher Forcing an related MLE-based methods, the generated samples need not be close to target samples to train.

However, character- and word-based representations of language are typically discrete (e.g., represented as discrete ``tokens"), and GANs require a gradient that back-propagates from the discriminator through the generated samples to train.
Solutions have been proposed to address GANs with discrete variables which use policy gradients based on the discriminator score~\citep{hjelm2017bsgan}, and other methods even applied similar policy gradients for discrete sequences of language~\citep{yu2016poem, li2017adversarial}.
However, for reasons of which are covered in detail below, these methods provide a high-variance gradient for the generator of sequences, which may limit the model's ability to scale to realistic real-world sequences of long length.

\subsection{The Actor-Critic Framework}
The actor-critic framework is a widely used approach to reinforcement learning to address credit assignment issues associated with making sequential discrete decisions.
In this framework, an ``actor" models a policy of actions, and a ``critic" estimates the expected reward or \emph{value function}.

Let us assume that there is a state, $s_t \in \mathcal{S}$ which is used as input to a conditional multinomial distribution of policies with density $\pi_{\theta}(a \mid s_{t})$ over actions, $a: \mathcal{S} \rightarrow \mathcal{S}$.
At timestep $t$, an agent / actor samples an action, $a_t \sim \pi_{\theta}(a \mid s_{t})$, and performs the action which generates the new state $s_{t+1}$.
This procedure ultimately leads to a reward, $R = \sum_{t=1}^T r(a_t; s_t)$ (where in general, the total reward could be an accumulation of intermediate rewards), over the whole episode, $S = s_1, \dots s_T$, where $s_T$ is some terminal state.

The REINFORCE algorithm uses this long term reward by running the agent / actor for an episode, gathering rewards, then using them to make an update to $\pi_{\theta}(a \mid s_{t})$ through a policy gradient. 
However there are some issues with this approach, as there is no mechanism to decide which actions may have affected the long-term outcomes and thus it can be ambiguous how to distribute the reward among them. 
This is referred to as the \emph{credit assignment issue}. On top of this, the Reinforce algorithm also suffers from high variance.

In actor-critic methods, the critic attempts to estimate the expected reward associated with each state, typically by using the Temporal Difference~\citep[TD,][]{sutton1988learning} learning.
The advantage is of actor-critic is, despite the added bias from estimating the expected reward, we can update the parameters of agent / actor at each step using the critic's estimates. This lowers the variance of gradient estimates, as the policy gradient is based on each state, $s_t$, rather than the whole episode, $S$. This also ensures faster convergence of the policy, as the updates can be made on-line as the episode, $S$, is generated.

\section{Actor-Critic under Adversarial Learning (ACtuAL)}
At this point, the analogy between the task of training a generative RNN adversarially and the reinforcement learning setting above should be clear.
Namely, the reward signal is provided by the discriminator, which corresponds more or less to the likelihood ratio between the distribution of generated sequences and the target distribution as it's trained to minimize the mis-classification rate.
The state, $s_t$, is a sequence of generated tokens, $\hat{Y}_{1, t} := \hat{y}_1, \dots, \hat{y}_{t-1}$, up to time $t$.
The actor is the generative RNN, whose policy, $\pi_{\theta}(a | s_t) = q_{\theta}(y_t | \hat{Y}_{1, t-1})$, corresponds to the conditional  distribution defined by the RNN.
In order to do better credit-assignment, we introduce a third component to the generative adversarial framework: the critic, which is optimized to estimate the expected reward at each step of generation.

\begin{figure*}[t]
\centering
\begin{tikzpicture}[
    black!50, text=black,
    font=\small	,
    node distance=2mm,
    dnode/.style={
        align=center,
        rectangle,minimum size=7mm,rounded corners,
        inner sep=5pt},
    rnode/.style={
        align=center,
        rectangle,
        minimum width=40mm,
        minimum height=7mm,
        rounded corners,
        inner sep=3pt,
        very thick,draw=black!50},
    tuplenode/.style={
        align=center,
        rectangle,minimum size=7mm,rounded corners,
        inner sep=5pt},
    darrow/.style={
        rounded corners,-latex,shorten <=5pt,shorten >=1pt,line width=2mm},
    mega thick/.style={line width=2pt}]
    
\matrix[row sep=12mm,column sep=15mm] {
    \node (critic) [rnode] {Critic}; &
    \node (obs) [rnode,fill=greenfill] {Observed Sequences};\\
    
    \node (actor) [rnode] {Recurrent Generator}; &
    \node (disc) [rnode,bottom color=redfill,top color=greenfill] {Discriminator};\\

    \node (env) [rnode] {Discrete Sampled Actions}; &
    \node (gen) [rnode,fill=redfill] {Generated Sequences};\\

\\
};

\begin{scope}[local bounding box=cycle_arrows]
    \draw[-latex,shorten <=1pt,shorten >=1pt,mega thick] (actor) to [bend left=45] (env);
    \draw[-latex,shorten <=1pt,shorten >=1pt,mega thick] (env) to [bend left=45] (actor);
\end{scope}

\draw[-latex,shorten <=1pt,shorten >=1pt,mega thick] (env) to [bend left=0] (gen);

\draw[-latex,shorten <=4pt,shorten >=1pt,mega thick] (actor) to [bend left=0] (critic);

\draw[-latex,shorten <=1pt,shorten >=1pt,mega thick,blue!50] (obs) to [bend left=0] (disc);

\draw[-latex,shorten <=1pt,shorten >=1pt,mega thick,red!50] (gen) to [bend left=0] (disc);

\draw[-latex,shorten <=1pt,shorten >=1pt,mega thick,dotted]  (disc) to [bend left = 8] (critic);

\node [below right=of critic] {Reward signal, $r_T$};

\begin{pgfonlayer}{background}
    \node (cycle) [fit=(actor) (env),rnode,inner sep=5pt,draw=moreredfill,mega thick,dotted] {};
\end{pgfonlayer}
\node [below right=2pt of cycle.south west,text=myred!80!white] {Actor.  t = 1 \dots T};
\end{tikzpicture}

\caption{Diagram illustrating the ACtuAL setup.  
An RNN (in this case with an autoregressive noise model) is sampled for N time steps, producing a generated sequence of N discrete tokens.  
A discriminator is trained as a binary classifier between the distribution of completed token sequences under this model and the true data distribution.  
A critic, using a TD(0) ~\citep{sutton1988learning} error and a sparse reward (only seeing the discriminator's score at the terminal step, $t = T$), estimates the expected reward or value function at each time step.  
In ACtuAL, the RNN generator (actor) is trained to maximize the value function of the critic for each time step.  
This is used as a policy gradient for the RNN generator at each time step.}
\label{fig:diagram}
\end{figure*}
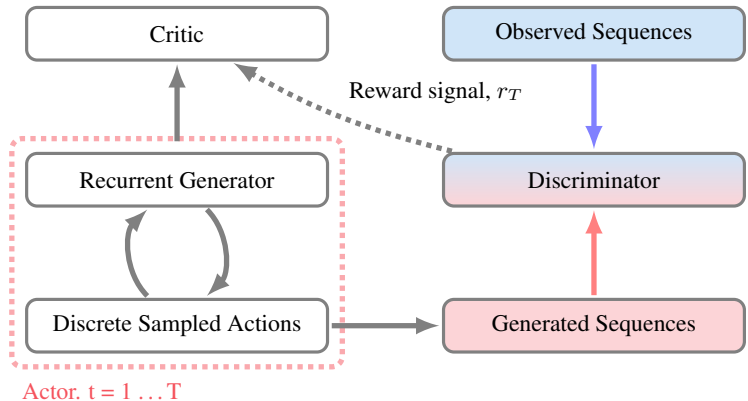

\subsection{ACtuAL for sequence generation}
Here we specify an ACtuAL architecture for natural language generation.
A diagram of the model is also shown in Figure~\ref{fig:diagram}. 

\textbf{The actor / generator}
The generator is modeled by a free-running RNN with parameters $\theta$ that generates each token conditioned on all previously generated tokens.
The actions are the set of all possible tokens in a vocabulary, $a := y \in \mathcal{V}$.
Given a generated sequence, $\hat{Y}_{1, t-1} = \hat{y}_1, \dots, \hat{y}_{t-1}$, the RNN defines a conditional multinomial distribution at time $t$, $q_{\theta}(y_t | \hat{Y}_{1, t-1})$, which modeled by a softmax layer on the outputs from the generator network.
The initial state is set to the $BOS$ token, specifying the beginning of the sequence, while the recurrent hidden state is initialized to $0$. 
At the terminal step, the generator has produced the terminal state corresponding to its set of actions or tokens, $s_T = a_1, \dots, a_T = \hat{y}_1, \dots, \hat{y}_T := \hat{Y}_{1, T}$.

\textbf{The discriminator}
The discriminator $D_\phi(Y)$\footnote{here we denote $Y$ as a full and terminating sequence of undetermined length.} is modeled as a bidirectional RNN~\citep{graves2012supervised} and is trained as a classifier between the generated sequences from $q_{\theta}$ and empirical target distribution, $p$, as in Equation~\ref{eq:gan}. 
The discriminator outputs a single scalar which is used as the reward, $r_T$, and can be interpreted as a scoring that the sample resembles the original data distribution. 

\textbf{The critic}
The critic, $\hat{Q}_{\psi}(a; \hat{T}_{1,t})$, is an RNN with parameters $\psi$, which is used to evaluate the stochastic policy of the generator based on expected reward.

In general actor-critic methods, given the current state, $\hat{Y}_{1 \dots t}$, we define a \emph{value function}:
\begin{equation}
V_{\theta}(\hat{Y}_{1, t}) = \EE_{\hat{Y}_{t+1, T} \sim p(\hat{Y}_{t+1, T} \mid \hat{Y}_{1, t})} \left[\sum_{\tau=t+1}^{T} r_{\tau}(\hat{y}_{\tau}; \hat{Y}_{1, t})\right].
\end{equation}
To estimate the value function, we define the value of an action $a \in \mathcal{V}$ as:
\begin{align}
Q(a; \hat{Y}_{1, t-1}) = \EE_{\hat{Y}_{t+1, T} \sim p(\hat{Y}_{t+1, T} \mid \hat{Y}_{1, t-1}, a)} \left[r_{t+1}(a; \hat{Y}_{1, t}) + \sum_{\tau=t+1}^{T} r_{\tau}(\hat{y}_{\tau}; \hat{Y}_{1, t-1}, a, \hat{Y}_{t+1, \tau}) \right]
\end{align}
When there is exactly one reward, e.g., when the sequence is only evaluated using the discriminator at the terminal state, $s_T$, all intermediate the value simplifies to:
\begin{align}
Q(a; \hat{Y}_{1, t-1}) = \EE_{\hat{Y}_{t+1, T} \sim p(\hat{Y}_{t+1, T} \mid \hat{Y}_{1, t-1}, a)} \left[D_{\phi}(\hat{Y}) \right].
\end{align}

\subsection{Algorithm details}

\begin{algorithm}[t]
    \begin{algorithmic}
        \State $(\theta, \phi, \psi)$
        \Comment{initialize the parameters of the generator, discriminator, and critic RNNs}
        \State $(\theta', \psi') \gets (\theta, \psi)$ 
        \Comment{initialize the parameters of the delayed generator and ``target" critic RNNs}
        \\
        \Repeat
            \State 1) \underline{\emph{Train the discriminator}}:
            \State $Y \sim p(Y)$
            \Comment{Draw $N$ sequences from the empirical target distribution}
            \State $\hat{Y} \sim q_{\theta}(Y)$
            \Comment{Draw $N$ samples from generator RNN}
            \State $T_{\phi}(Y, \hat{Y}) \gets \frac{1}{N} \sum_n \log{D_{\phi}(Y^{(n)})} - \frac{1}{N} \sum_n \log{D_{\phi}(\hat{Y}^{(n)})}$
            \\
            \Comment{Estimate the discriminator objective}
            \State $\phi \gets \phi + \gamma_D \nabla_{\phi} T_{\phi}(Y, \hat{Y})$
            \Comment{Update the discriminator parameters}
            \\
            \State 2) \underline{\emph{Train the actor (generator) / critic}}:
            \For{mini-batch}
            \State $\hat{Y}' \sim q_{\theta'}(Y)$
            \Comment{Draw a sample from delayed generator RNN}
            \State $\kappa_t \gets \sum_{a \in \mathcal{V}} q_{\theta'}(a \mid \hat{Y}_{1,t}) \hat{Q}_{\psi'} (a; \hat{Y}_{1,t})$
            \Comment{Compute targets for the critic}
            \State  $C_t \gets \sum_{a}\left(\hat{Q}_{\phi}(a; \hat{Y}_{1, t-1}) - \frac{1}{|\mathcal{V}|}\sum_{b \in \mathcal{V}}\hat{Q}_{\psi}(b; \hat{Y}_{1, t-1}) \right)^2$
            \\
            \Comment{Compute the variance penalties}
            \State $\psi \gets \psi - \gamma_Q \nabla_{\psi} \left( \sum_{t=1}^{T} (\hat{Q}_{\psi}(\hat{y}_t; \hat{Y}_{1, t-1}) - \kappa_t)^2 + \lambda  C_t \right)$
            \Comment{Update the critic parameters}
            \State $V \gets \sum_{t=1}^{T} \sum_{a \in \mathcal{V}} \nabla_{\theta} q_{\theta}(a \mid \hat{Y}_{1,t-1}) \hat{Q}_{\psi}(a; \hat{Y}_{1,t-1})$
            \Comment{Compute the expected return}
            \State $\theta \gets \theta + \gamma_D \nabla_{\theta} V$
            \Comment{Update the generator parameters}
            \State $\theta' \gets \tau \theta + (1 - \tau) \theta'$
            \State $\phi' \gets \tau \phi + (1 - \tau) \phi'$
            \Comment{Update the delayed actor and target critic, with constant $\tau \ll 1$}
            \EndFor
            \\
    \Until{convergence}
    \end{algorithmic}
\caption{\label{al:acgan}. ACtuAL: Actor-Critic under Adversarial Learning}
\end{algorithm}

Here we describe the training objectives for ACtuAL.
The complete procedure is provided in Algorithm~\ref{al:acgan}, though we will cover specific implementation details here.

\paragraph{Training the discriminator}
Samples from the generated distribution, $q_{\theta}$, are first drawn according to the generative RNN. 
Next, these samples along with samples from the target distribution are used to update the discriminator according to Equation~\ref{eq:gan}.
The outputs of the discriminator are then used as a reward signal for training both the critic and the generator.

\paragraph{Reducing variance using Temporal-Difference TD(0) learning:}
\emph{Policy evaluation} involves calculating the expected return at time $t$, $\sum_{\tau=t}^T r_{\tau}(\hat{y}_{\tau}; \hat{Y}_{1, \tau - 1})$, by a Monte-Carlo roll-out, which are used as targets for the critic, $\hat{Q}_{\psi}(\hat{y}_t; \hat{Y}_{1, t - 1})$.
The parameters of the critic, $\psi$, are typically updated to minimize the square-error between the target value and this expected return.
However, this Monte-Carlo roll out leads to very high variance which grows with the number
of steps $T$.
In order to reduce the variance, but at the risk of increasing bias, we use the Temporal Difference (TD) method (in our experiments, TD(0)) for policy evaluation~\citep{sutton1988learning}. 
Specifically, we use the right-hand side of the Bellman equation, $\kappa_t = r_t(\hat{y}_t;\hat{Y}_{1, t - 1}) + \sum_{a \in \mathcal{A}}
p(a|\hat{Y}_{1 \ldots t}) \hat{Q}_{\psi}(a;\hat{Y}_{1, t})$, as the
target for $\hat{Q}_{\psi}(\hat{y}_t; \hat{Y}_{1, t - 1})$. In our experiments, we don't back-propagate through the right hand side of
Bellman equation, i.e., we treat the right hand side as a fixed quantity. 

\paragraph{Training the generator}
The policy gradient for the generator at each step is estimated using the output of the critic at each time step.
However, as the true reward is provided by the discriminator at the terminal time step, $T$, we use the discriminator output, $D_{\phi}(\hat{Y}_{1,T})$ for the policy gradient at the final step.

\paragraph{Delayed actor and target critic}
In actor-critic methods, the actor and critic use each other's output for training, which can potentially cause instabilities during training. 
Hence, the model could spend time catching up to changes in the other, which could result in a poor / highly variable training signal. 
Following the work of \citet{lillicrap2015continuous}, it is common practice to use an additional \emph{target critic} and \emph{delayed actor} that are updated less frequently. 
When training the critic, the actions are drawn from the delayed actor, and its targets are provided by the delayed actor and target critic. 
We update the parameters of the target critic and delayed actor by linearly interpolating their parameters with those of the trained critic and actor, respectively.

\paragraph{Large action spaces:}

When training word-level generative models, the number of actions can be in the tens of thousands.  This means that many rarely seen words are almost never sampled and thus the critic's value estimate for these words could have very high variance.  To address this we add a penalty, $C_t$, at every step which encouraged the critic's outputs for different action to be close to the critic's mean value:  
\begin{align} 
    C_t = \sum_{a}\left( 
        \hat{Q}(a; \hat{Y}_{1\ldots t-1}) 
        - \frac{1}{|\mathcal{V}|}\sum_{b}\hat{Q}(b; \hat{Y}_{1 \ldots t-1}) 
    \right)^2 \label{eq:variance_penalty}
\end{align} 
This same penalty was previously employed in the context of learning simple algorithms with $Q$-learning~\citep{zaremba2015learning, bahdanau2016actorcritic}.  Experimentally we found that this penalty was essential for successful training of word-level as well as character-level models.  

\paragraph{Pre-training the actor and critic}
Before using Algorithm~\ref{al:acgan}, instead of starting with the actor and critic with random parameters, we first pre-train the actor network using Teacher Forcing. 
Next, we pre-train critic network, keeping the actor network fixed (also using the discriminator, which is trained). 
We found that pre-training actor and critic networks were essential, otherwise the learning signal for actor network (or policy) was not good enough to learn anything meaningful. In all the experiments, we consider adding weighted log-likelihood gradient in addition to actor's gradient estimate. 

\section{Related Work}
\paragraph{Actor-critic in GANs}
The connection between actor-critic and adversarial learning was explored first in ~\citet{pfau2016connecting}.
In our work, we explore this connection more thoroughly and empirically in the difficult setting of natural language generation.

\paragraph{SeqGAN}
SeqGAN~\citep{yu2016poem} uses Monte-Carlo search to perform credit assignment through the discrete sampling process. 
Monte-Carlo search is a reinforcement learning (RL) technique that assigns credit through Monte-Carlo roll-out. 
Though this technique is unbiased, gradients can have high variance, and therefore learning takes long to converge.
This significantly limits the model's to scale to large real-world sequences.

\paragraph{Discrete GANs trained on continuous spaces}
In the Gradient Penalty Wasserstein GAN~\citep[WGAN-GP,][]{gulrajani2017improved}, the authors approach the task of generating text adversarially by using a 1D-CNN to generate sequences of a fixed length.  
In this work, the discreteness of the data was ignored and the GAN was trained directly on the continuous softmax outputs.
This usually should prevent GANs from working, as it should be relatively easy for the discriminator to differentiate between continuous-fake and discrete-real sequences.
However, the Wasserstein distance is a ``weaker" metric than the JSD~\citep{arjovsky2017wgan}, or at least the regularization necessarily makes it so, so a GAN trained on this way can provide a meaningful gradient even in this setting. 
\citet{rajeswar2017adversarial} follows this approach, using the softmax probabilities with both recurrent and convolutional architectures on both word- and character-level language modeling tasks.
While these approaches demonstrate significant progress in training GANs on discrete and sequential data, it is not clear the consequences of the continuous approximation on the resulting generator.

Several other works avoid direct discrete sequence generation by working directly with continuous spaces.
In Professor Forcing~\citep{goyalprofessor}, the authors consider training the discriminator on continuous hidden states of an RNN to improve long-term dependencies while bypassing the challenges in using adversarial learning with discrete states.
\citet{kim2017adversarially} train the generator to match the intermediate bottleneck state with an autoencoder trained to reconstruct discrete sequences.
While worthwhile and interesting approaches, they represent an orthogonal direction to ours, none address the problems associated with credit assignment across long sequences.

\paragraph{GANs trained on policy gradients}
\citet{li2017adversarial} explored using the REINFORCE algorithm~\citep{williams1992reinforce} along with a discriminator for the task of dialogue generation. 
As mentioned above, REINFORCE does not scale well to real-world sequences, such as with word-level language generation, due to the associated high-variance of estimates. 

Boundary-seeking GANs~\citep[BGAN,][]{hjelm2017bsgan} uses a policy gradient derived directly from an estimate of the likelihood ratio. 
They provide two learning algorithms, one that uses normalized importance weights and the other which resembles REINFORCE with a baseline.
The authors used a similar CNN-based approach as \citet{gulrajani2017improved} and achieved similar results.
Their work is not incompatible to ours, as the ``weights" could be used as the reward signal in our actor-critic framework.

\paragraph{Actor-Critic with pre-defined rewards}
Additionally, actor-critic and other reinforcement learning algorithms have been used to improve generative models for discrete sequences.  \citet{bahdanau2016actorcritic} proposed using an actor-critic setup for generating text, using BLEU score~\citep{bleu} and character level error rate as the rewards.  Optimizing for BLEU score using a variant of REINFORCE was also explored in \citet{ranzato2015sequence}.  

\section{Experiments}
We evaluated Actor-Critic under Adversarial Learning (ACtuAL) on several competitive language modeling benchmarks: word- and character-level Penn-Treebank datasets~\citep{marcus1993building}, the Chinese Poetry dataset~\citep{zhang2014chinese}, and the Text8 dataset~\citep{mahoney2011large}.  
The main goal of these experiments is to show that training in the actor-critic framework helps to make the generator distribution closer to the true data distribution.  

Even though the generative adversarial network is not optimized for likelihood, surprisingly we empirically observe that training with the ACtuAL objective achieves a better likelihood than when training with only the MLE objective.  

We evaluate our model on both word- and character-level language modeling datasets.
Word-level language modeling contains a much larger output (action) space compared to character-level language modeling. 
For all our experiments, we first pre-trained the actor / generator with using MLE and then pretrained the critic with the actor parameters fixed. 
We then fined tuned all networks (actor / generator, critic, and discriminator) using Algorithm~\ref{al:acgan}.
Best results for all the experiments were obtained by including the weighted log-likelihood gradient (weighted by 0.1 in all experiments).

\subsection{Word Level Experiments}

We investigated the ability of ACtuAL to improve language modeling at the word level.  Word level language modeling is a challenge because the vocabulary is relatively large (usually tens of thousands of possible words), meaning that the space being modeled is high-dimensional and very sparse.  

\paragraph{Word-level Penn Treebank}
The Penn-Treebank dataset has a vocabulary of $10,000$ unique words.  
The training set contains $930,000$ words, the validation set contains $74,000$ words and the test set contains $82,000$ words. 
We divide the training set into non-overlapping sequences each with length of $30$. 
We monitor the negative log-likelihood (NLL) on both the training set and the validation set, and used NLL on the validation set as an early stopping criteria.  
We found that using ACtuAL improved the NLL over the Teacher Forcing baseline, and that our best results were achieved while using the weighted log-likelihood gradient (weighted by 0.1) and using the variance penalty hyper-parameter of $\lambda = 5$.

Our actor network / generator and critic networks were both single-layer LSTMs \cite{hochreiter1997long} with $1024$ hidden units. 
The discriminator is a single-layer GRU \cite{chung2014empirical} with $512$ units. 
We use the Adam optimizer~\citep{kingma2014adam} with a learning rate of $0.0001$. We also clip the gradients to have maximum norm equal of $10$. 

\begin{table}[!htb]
\centering
\label{ptb-word}
\begin{tabular}{|Sl|Sl|Sl|Sl}
\cline{1-3}
Method &   Valid & Test  &  \\ \cline{1-3}
Teacher Forcing &  154.387 & 152.854 &   \\ \cline{1-3}

\shortstack[l]{Teacher Forcing + Actor Critic} & \textbf{150.784} & \textbf{149.365} &\\ \cline{1-3}
\end{tabular}

\caption{Negative Log-Likelihood (NLL) Results on word-level Penn Treebank.  }

\end{table}

\paragraph{Chinese Poetry Dataset}
The Chinese Poetry dataset consists of Chinese poems that were used to evaluate adversarial training methods for natural language in \citet{yu2016seq} and \citet{che2017poem}. 
Each poem in the dataset has $4$-lines and a variable number of characters in each line. 
We use lines of length $5$ (poem-$5$) and $7$ (poem-$7$) with the train/validation/test splits used in \citet{che2017poem}. 
We evaluate both likelihood and qualitative sample quality for this task. 

\begin{table}[!htb]
\centering
\label{chinese-poem}
\begin{tabular}{|Sl|Sl|Sl|Sl|Sl}
\cline{1-4}
Method & Train  & Valid & Test  &  \\ \cline{1-4}
Teacher Forcing & 29.79  & 32.39 & 32.15    \\   \cline{1-4}
\shortstack[l]{Teacher Forcing  + Actor Critic} & 32.72 & \textbf{32.05} & \textbf{31.83} &   \\ \cline{1-4}
\end{tabular}

\caption{Negative Log-Likelihood (NLL) Results on the Chinese poetry dataset.  }

\end{table}

\subsection{Character Level Experiments}
We investigated the ability of ACtuAL to improve character-level language modeling.  
Our motivation in exploring character-level language modeling is that these sequences have longer-term dependencies than word-level language modeling, suggesting that there may be some benefit in incorporating the critic into the GAN objective.  
Benefits from applying adversarial training to character-level language modeling had previously been observed in \citet{chung2016character}.

\paragraph{Character-Level Penn TreeBank }
We follow the partition in \citet{mikolov2012subword} and divide the Penn TreeBank dataset into train/validation/test sets. We segment the data into length of $100$ non-overlapping sequences. 

We train single-layer GRUs with $1000$ units on the character-level PTB dataset. 
We train the GRUs with a learning rate of 0.0001, optimizing using Adam, and clip gradients with 
threshold $1$.  
We found that using ACtuAL improved BPC over the Teacher Forcing baseline, and that our best results were achieved while using the weighted log-likelihood gradient (weighted by 0.1) as well as using the penalty term $\lambda = 0.1$.

\begin{table}[!htb]
\centering
\label{my-label}
\begin{tabular}{|Sl|Sl|Sl|Sl}
\cline{1-3}
Method &   Valid & Test  &  \\ \cline{1-3}
Teacher Forcing &  1.47 &  1.38 &   \\ \cline{1-3}
\shortstack[l]{Teacher Forcing  + Actor Critic} &  \textbf{ 1.43} & \textbf{1.34} &   \\ \cline{1-3}
\end{tabular}

\caption{Bit per character (bpc) Results on char-level Penn Treebank.  }

\end{table}

\subsection{Text8}
Here, we would evaluate our model on a very large natural language dataset to show scalability properties.
The Text8 dataset is a clean version of Enwik8 dataset, which is derived from Wikipedia articles. 
Text8 contains the first 100M characters from Wikipedia, and the characters are restricted to alphabets and spaces only. The Text8 dataset is derived from Wikepedia as well, and it is the first 100M characters in the Wikipedia articles.

Following the setup in \citet{mikolov2012subword,zhang2016architectural}, we use the first 90M characters for training, the next 5M for validation, and the final 5M characters for testing. 
We train on non-overlapping sequences of length $180$. 
We use the same architecture for the generator as \citet{cooijmans2016recurrent} and trained a single-layer LSTM with $2000$ units on the character-level Text8 dataset. 
The critic is a Gated Recurrent Unit (GRU) \cite{chung2014empirical} with 2000 units. The discriminator is a GRU with $512$ units. We used the Adam with a learning rate of 0.0001 for training of all models. 
We found that ACtuAL improved BPC over the teacher forcing baseline.

\begin{table}[!htb]
\centering
\label{text8-cha}
\begin{tabular}{|Sl|Sl|Sl|Sl}
\cline{1-3}
Method &   Valid & Test  &  \\ \cline{1-3}
Teacher Forcing &  1.42 &  1.41 &   \\ \cline{1-3}
\shortstack[l]{Teacher Forcing  + Actor Critic} &  \textbf{1.40 } & \textbf{1.39} &   \\ \cline{1-3}
\end{tabular}

\caption{Bit per character (bpc) Results on char-level Text8.  }
\end{table}

\section{Conclusion}

Modeling discrete sequences is a central task in many fields, including language modeling, machine translation, music synthesis, and forecasting.  
Generative Adversarial Networks are a state of the art generative model for image data, yet they require passing gradients from the discriminator to the generator, which cannot be done exactly when the data is discrete.  
We proposed to palliate this limitation with the Actor-Critic under Adversarial Learning (ACtuAL) architecture, which trains a critic to predict the ability of a state, probability pair at a given timestep to fool the discriminator over future timesteps.  This avoids the back-propagation through discrete states issue.  
On the Penn-Treebank language modeling task we showed that this actor-critic architecture can learn to reduce the accuracy of the discriminator while at the same time improving the likelihood of the generator.
Finally, we believe that ACtuAL can learn to interact with an external environment (for example, a knowledge graph) to improve its predictions, and we leave this for future work.  

\section{Acknowledgements}

This work was done as a course project for Professor Doina Precup's Reinforment Learning course taught at McGill University. The authors would like to thank Pierre-Luc Bacon, Dzmitry Bahdanau,  Krishna Kumar, Aravind Srinivas, Peter Henderson, Ahmed Touati, Prasanna Parthasarathi for useful feedback and discussions  as well as NSERC, CIFAR, Google, Samsung, Nuance, IBM and Canada Research Chairs for funding, and Compute Canada and NVIDIA for computing resources. RDH would like to thank IVADO for their support. The authors would also like to express debt of gratitude towards those who contributed to Theano over the years (now that it is being sunset), for making it such a great tool.

\bibliography{iclr2018_conference}

\begin{thebibliography}{45}
\providecommand{\natexlab}[1]{#1}
\providecommand{\url}[1]{\texttt{#1}}
\expandafter\ifx\csname urlstyle\endcsname\relax
  \providecommand{\doi}[1]{doi: #1}\else
  \providecommand{\doi}{doi: \begingroup \urlstyle{rm}\Url}\fi

\bibitem[{Arjovsky} et~al.(2017){Arjovsky}, {Chintala}, and
  {Bottou}]{arjovsky2017wgan}
M.~{Arjovsky}, S.~{Chintala}, and L.~{Bottou}.
\newblock {Wasserstein GAN}.
\newblock \emph{ArXiv e-prints}, January 2017.

\bibitem[Bahdanau et~al.(2016)Bahdanau, Brakel, Xu, Goyal, Lowe, Pineau,
  Courville, and Bengio]{bahdanau2016actorcritic}
Dzmitry Bahdanau, Philemon Brakel, Kelvin Xu, Anirudh Goyal, Ryan Lowe, Joelle
  Pineau, Aaron~C. Courville, and Yoshua Bengio.
\newblock An actor-critic algorithm for sequence prediction.
\newblock \emph{CoRR}, abs/1607.07086, 2016.
\newblock URL \url{http://arxiv.org/abs/1607.07086}.

\bibitem[Bengio et~al.(2015)Bengio, Vinyals, Jaitly, and
  Shazeer]{bengio2015scheduled}
Samy Bengio, Oriol Vinyals, Navdeep Jaitly, and Noam Shazeer.
\newblock Scheduled sampling for sequence prediction with recurrent neural
  networks.
\newblock In \emph{Advances in Neural Information Processing Systems}, pp.\
  1171--1179, 2015.

\bibitem[Bengio et~al.(2013)Bengio, L{\'e}onard, and
  Courville]{bengio2013estimating}
Yoshua Bengio, Nicholas L{\'e}onard, and Aaron Courville.
\newblock Estimating or propagating gradients through stochastic neurons for
  conditional computation.
\newblock \emph{arXiv preprint arXiv:1308.3432}, 2013.

\bibitem[Che et~al.(2017)Che, Li, Zhang, Hjelm, Li, Song, and
  Bengio]{che2017poem}
Tong Che, Yanran Li, Ruixiang Zhang, R.~Devon Hjelm, Wenjie Li, Yangqiu Song,
  and Yoshua Bengio.
\newblock Maximum-likelihood augmented discrete generative adversarial
  networks.
\newblock \emph{CoRR}, abs/1702.07983, 2017.
\newblock URL \url{http://arxiv.org/abs/1702.07983}.

\bibitem[Cho et~al.(2014)Cho, Van~Merri{\"e}nboer, Gulcehre, Bahdanau,
  Bougares, Schwenk, and Bengio]{cho2014learning}
Kyunghyun Cho, Bart Van~Merri{\"e}nboer, Caglar Gulcehre, Dzmitry Bahdanau,
  Fethi Bougares, Holger Schwenk, and Yoshua Bengio.
\newblock Learning phrase representations using rnn encoder-decoder for
  statistical machine translation.
\newblock \emph{arXiv preprint arXiv:1406.1078}, 2014.

\bibitem[Chorowski et~al.(2015)Chorowski, Bahdanau, Serdyuk, Cho, and
  Bengio]{chorowski2015attention}
Jan Chorowski, Dzmitry Bahdanau, Dmitriy Serdyuk, KyungHyun Cho, and Yoshua
  Bengio.
\newblock Attention-based models for speech recognition.
\newblock \emph{CoRR}, abs/1506.07503, 2015.
\newblock URL \url{http://arxiv.org/abs/1506.07503}.

\bibitem[Chung et~al.(2014)Chung, Gulcehre, Cho, and
  Bengio]{chung2014empirical}
Junyoung Chung, Caglar Gulcehre, KyungHyun Cho, and Yoshua Bengio.
\newblock Empirical evaluation of gated recurrent neural networks on sequence
  modeling.
\newblock \emph{arXiv preprint arXiv:1412.3555}, 2014.

\bibitem[Chung et~al.(2016)Chung, Cho, and Bengio]{chung2016character}
Junyoung Chung, Kyunghyun Cho, and Yoshua Bengio.
\newblock A character-level decoder without explicit segmentation for neural
  machine translation.
\newblock \emph{arXiv preprint arXiv:1603.06147}, 2016.

\bibitem[Cooijmans et~al.(2016)Cooijmans, Ballas, Laurent, G{\"u}l{\c{c}}ehre,
  and Courville]{cooijmans2016recurrent}
Tim Cooijmans, Nicolas Ballas, C{\'e}sar Laurent, {\c{C}}a{\u{g}}lar
  G{\"u}l{\c{c}}ehre, and Aaron Courville.
\newblock Recurrent batch normalization.
\newblock \emph{arXiv preprint arXiv:1603.09025}, 2016.

\bibitem[Flunkert et~al.(2017)Flunkert, Salinas, and
  Gasthaus]{flunkert2017deepar}
Valentin Flunkert, David Salinas, and Jan Gasthaus.
\newblock Deepar: Probabilistic forecasting with autoregressive recurrent
  networks.
\newblock \emph{arXiv preprint arXiv:1704.04110}, 2017.

\bibitem[Goodfellow(2016)]{goodfellow2016nips}
Ian Goodfellow.
\newblock Nips 2016 tutorial: Generative adversarial networks.
\newblock \emph{arXiv preprint arXiv:1701.00160}, 2016.

\bibitem[Goodfellow et~al.(2014)Goodfellow, Pouget-Abadie, Mirza, Xu,
  Warde-Farley, Ozair, Courville, and Bengio]{goodfellow2014generative}
Ian Goodfellow, Jean Pouget-Abadie, Mehdi Mirza, Bing Xu, David Warde-Farley,
  Sherjil Ozair, Aaron Courville, and Yoshua Bengio.
\newblock Generative adversarial nets.
\newblock In \emph{Advances in neural information processing systems}, pp.\
  2672--2680, 2014.

\bibitem[Goyal et~al.(2016)Goyal, Lamb, Zhang, Zhang, Courville, and
  Bengio]{goyalprofessor}
Anirudh Goyal, Alex Lamb, Ying Zhang, Saizheng Zhang, Aaron Courville, and
  Yoshua Bengio.
\newblock Professor forcing: A new algorithm for training recurrent nets.
\newblock In \emph{Advances In Neural Information Processing Systems}, pp.\
  4601--4609, 2016.

\bibitem[Graves et~al.(2012)]{graves2012supervised}
Alex Graves et~al.
\newblock \emph{Supervised sequence labelling with recurrent neural networks},
  volume 385.
\newblock Springer, 2012.

\bibitem[Gu et~al.(2015)Gu, Levine, Sutskever, and Mnih]{gu2015muprop}
Shixiang Gu, Sergey Levine, Ilya Sutskever, and Andriy Mnih.
\newblock Muprop: Unbiased backpropagation for stochastic neural networks.
\newblock \emph{arXiv preprint arXiv:1511.05176}, 2015.

\bibitem[Gulrajani et~al.(2017)Gulrajani, Ahmed, Arjovsky, Dumoulin, and
  Courville]{gulrajani2017improved}
Ishaan Gulrajani, Faruk Ahmed, Mart{\'{\i}}n Arjovsky, Vincent Dumoulin, and
  Aaron~C. Courville.
\newblock Improved training of wasserstein gans.
\newblock \emph{CoRR}, abs/1704.00028, 2017.
\newblock URL \url{http://arxiv.org/abs/1704.00028}.

\bibitem[{Hjelm} et~al.(2017){Hjelm}, {Jacob}, {Che}, {Cho}, and
  {Bengio}]{hjelm2017bsgan}
R~Devon {Hjelm}, Athul~Paul. {Jacob}, Tong {Che}, Kyunghyun {Cho}, and Yoshua
  {Bengio}.
\newblock {Boundary-Seeking Generative Adversarial Networks}.
\newblock \emph{ArXiv e-prints}, February 2017.

\bibitem[Hochreiter \& Schmidhuber(1997)Hochreiter and
  Schmidhuber]{hochreiter1997long}
Sepp Hochreiter and J{\"u}rgen Schmidhuber.
\newblock Long short-term memory.
\newblock \emph{Neural computation}, 9\penalty0 (8):\penalty0 1735--1780, 1997.

\bibitem[{Jang} et~al.(2016){Jang}, {Gu}, and {Poole}]{jang2016categorical}
E.~{Jang}, S.~{Gu}, and B.~{Poole}.
\newblock {Categorical Reparameterization with Gumbel-Softmax}.
\newblock \emph{ArXiv e-prints}, November 2016.

\bibitem[Kim et~al.(2017)Kim, Zhang, Rush, LeCun, et~al.]{kim2017adversarially}
Yoon Kim, Kelly Zhang, Alexander~M Rush, Yann LeCun, et~al.
\newblock Adversarially regularized autoencoders for generating discrete
  structures.
\newblock \emph{arXiv preprint arXiv:1706.04223}, 2017.

\bibitem[Kingma \& Ba(2014)Kingma and Ba]{kingma2014adam}
Diederik Kingma and Jimmy Ba.
\newblock Adam: A method for stochastic optimization.
\newblock \emph{arXiv preprint arXiv:1412.6980}, 2014.

\bibitem[Li et~al.(2017)Li, Monroe, Shi, Ritter, and
  Jurafsky]{li2017adversarial}
Jiwei Li, Will Monroe, Tianlin Shi, Alan Ritter, and Dan Jurafsky.
\newblock Adversarial learning for neural dialogue generation.
\newblock \emph{CoRR}, abs/1701.06547, 2017.
\newblock URL \url{http://arxiv.org/abs/1701.06547}.

\bibitem[Lillicrap et~al.(2015)Lillicrap, Hunt, Pritzel, Heess, Erez, Tassa,
  Silver, and Wierstra]{lillicrap2015continuous}
Timothy~P Lillicrap, Jonathan~J Hunt, Alexander Pritzel, Nicolas Heess, Tom
  Erez, Yuval Tassa, David Silver, and Daan Wierstra.
\newblock Continuous control with deep reinforcement learning.
\newblock \emph{arXiv preprint arXiv:1509.02971}, 2015.

\bibitem[Maddison et~al.(2016)Maddison, Mnih, and Teh]{maddison2016concrete}
Chris~J Maddison, Andriy Mnih, and Yee~Whye Teh.
\newblock The concrete distribution: A continuous relaxation of discrete random
  variables.
\newblock \emph{arXiv preprint arXiv:1611.00712}, 2016.

\bibitem[Mahoney(2011)]{mahoney2011large}
Matt Mahoney.
\newblock Large text compression benchmark, 2011.

\bibitem[Marcus et~al.(1993)Marcus, Marcinkiewicz, and
  Santorini]{marcus1993building}
Mitchell~P Marcus, Mary~Ann Marcinkiewicz, and Beatrice Santorini.
\newblock Building a large annotated corpus of english: The penn treebank.
\newblock \emph{Computational linguistics}, 19\penalty0 (2):\penalty0 313--330,
  1993.

\bibitem[Mikolov \& Zweig(2012)Mikolov and Zweig]{mikolov2012context}
Tomas Mikolov and Geoffrey Zweig.
\newblock Context dependent recurrent neural network language model.
\newblock \emph{SLT}, 12:\penalty0 234--239, 2012.

\bibitem[Mikolov et~al.(2012)Mikolov, Sutskever, Deoras, Le, Kombrink, and
  Cernocky]{mikolov2012subword}
Tom{\'a}{\v{s}} Mikolov, Ilya Sutskever, Anoop Deoras, Hai-Son Le, Stefan
  Kombrink, and Jan Cernocky.
\newblock Subword language modeling with neural networks.
\newblock \emph{preprint (http://www. fit. vutbr. cz/imikolov/rnnlm/char.
  pdf)}, 2012.

\bibitem[Nowozin et~al.(2016)Nowozin, Cseke, and Tomioka]{nowozin2016f}
Sebastian Nowozin, Botond Cseke, and Ryota Tomioka.
\newblock f-gan: Training generative neural samplers using variational
  divergence minimization.
\newblock In \emph{Advances in Neural Information Processing Systems}, pp.\
  271--279, 2016.

\bibitem[Papineni et~al.(2002)Papineni, Roukos, Ward, and Zhu]{bleu}
Kishore Papineni, Salim Roukos, Todd Ward, and Wei-Jing Zhu.
\newblock Bleu: a method for automatic evaluation of machine translation.
\newblock In \emph{Proceedings of the 40th annual meeting on association for
  computational linguistics}, pp.\  311--318. Association for Computational
  Linguistics, 2002.

\bibitem[Pfau \& Vinyals(2016)Pfau and Vinyals]{pfau2016connecting}
David Pfau and Oriol Vinyals.
\newblock Connecting generative adversarial networks and actor-critic methods.
\newblock \emph{arXiv preprint arXiv:1610.01945}, 2016.

\bibitem[Radford et~al.(2015)Radford, Metz, and
  Chintala]{radford2015unsupervised}
Alec Radford, Luke Metz, and Soumith Chintala.
\newblock Unsupervised representation learning with deep convolutional
  generative adversarial networks.
\newblock \emph{arXiv preprint arXiv:1511.06434}, 2015.

\bibitem[Rajeswar et~al.(2017)Rajeswar, Subramanian, Dutil, Pal, and
  Courville]{rajeswar2017adversarial}
Sai Rajeswar, Sandeep Subramanian, Francis Dutil, Christopher~Joseph Pal, and
  Aaron~C. Courville.
\newblock Adversarial generation of natural language.
\newblock \emph{CoRR}, abs/1705.10929, 2017.
\newblock URL \url{http://arxiv.org/abs/1705.10929}.

\bibitem[Ranzato et~al.(2015)Ranzato, Chopra, Auli, and
  Zaremba]{ranzato2015sequence}
Marc'Aurelio Ranzato, Sumit Chopra, Michael Auli, and Wojciech Zaremba.
\newblock Sequence level training with recurrent neural networks.
\newblock \emph{arXiv preprint arXiv:1511.06732}, 2015.

\bibitem[Sutskever et~al.(2014)Sutskever, Vinyals, and
  Le]{sutskever2014sequence}
Ilya Sutskever, Oriol Vinyals, and Quoc~V. Le.
\newblock Sequence to sequence learning with neural networks.
\newblock In \emph{Advances in Neural Information Processing Systems 27: Annual
  Conference on Neural Information Processing Systems 2014, December 8-13 2014,
  Montreal, Quebec, Canada}, pp.\  3104--3112, 2014.

\bibitem[Sutton(1988)]{sutton1988learning}
Richard~S Sutton.
\newblock Learning to predict by the methods of temporal differences.
\newblock \emph{Machine learning}, 3\penalty0 (1):\penalty0 9--44, 1988.

\bibitem[Tucker et~al.(2017)Tucker, Mnih, Maddison, and
  Sohl-Dickstein]{tucker2017rebar}
George Tucker, Andriy Mnih, Chris~J Maddison, and Jascha Sohl-Dickstein.
\newblock Rebar: Low-variance, unbiased gradient estimates for discrete latent
  variable models.
\newblock \emph{arXiv preprint arXiv:1703.07370}, 2017.

\bibitem[Williams(1992)]{williams1992reinforce}
Ronald~J. Williams.
\newblock Simple statistical gradient-following algorithms for connectionist
  reinforcement learning.
\newblock \emph{Machine Learning}, 8\penalty0 (3):\penalty0 229--256, 1992.
\newblock ISSN 1573-0565.
\newblock \doi{10.1007/BF00992696}.
\newblock URL \url{http://dx.doi.org/10.1007/BF00992696}.

\bibitem[Williams \& Zipser(1989)Williams and Zipser]{williams1989learning}
Ronald~J Williams and David Zipser.
\newblock A learning algorithm for continually running fully recurrent neural
  networks.
\newblock \emph{Neural computation}, 1\penalty0 (2):\penalty0 270--280, 1989.

\bibitem[Yu et~al.(2016{\natexlab{a}})Yu, Zhang, Wang, and Yu]{yu2016poem}
Lantao Yu, Weinan Zhang, Jun Wang, and Yong Yu.
\newblock Seqgan: Sequence generative adversarial nets with policy gradient.
\newblock \emph{CoRR}, abs/1609.05473, 2016{\natexlab{a}}.
\newblock URL \url{http://arxiv.org/abs/1609.05473}.

\bibitem[Yu et~al.(2016{\natexlab{b}})Yu, Zhang, Wang, and Yu]{yu2016seq}
Lantao Yu, Weinan Zhang, Jun Wang, and Yong Yu.
\newblock Seqgan: Sequence generative adversarial nets with policy gradient.
\newblock \emph{CoRR}, abs/1609.05473, 2016{\natexlab{b}}.
\newblock URL \url{http://arxiv.org/abs/1609.05473}.

\bibitem[Zaremba et~al.(2015)Zaremba, Mikolov, Joulin, and
  Fergus]{zaremba2015learning}
Wojciech Zaremba, Tomas Mikolov, Armand Joulin, and Rob Fergus.
\newblock Learning simple algorithms from examples.
\newblock \emph{arXiv preprint arXiv:1511.07275}, 2015.

\bibitem[Zhang et~al.(2016)Zhang, Wu, Che, Lin, Memisevic, Salakhutdinov, and
  Bengio]{zhang2016architectural}
Saizheng Zhang, Yuhuai Wu, Tong Che, Zhouhan Lin, Roland Memisevic, Ruslan~R
  Salakhutdinov, and Yoshua Bengio.
\newblock Architectural complexity measures of recurrent neural networks.
\newblock In \emph{Advances in Neural Information Processing Systems}, pp.\
  1822--1830, 2016.

\bibitem[Zhang \& Lapata(2014)Zhang and Lapata]{zhang2014chinese}
Xingxing Zhang and Mirella Lapata.
\newblock Chinese poetry generation with recurrent neural networks.
\newblock In \emph{EMNLP}, pp.\  670--680, 2014.

\end{thebibliography}
\bibliographystyle{iclr2018_conference}

\end{document}